\documentclass{article}

\usepackage{microtype}
\usepackage{graphicx}
\usepackage{booktabs} 
\usepackage{multirow}
\usepackage{subfig}
\usepackage{lipsum}  
\usepackage{enumitem}
\usepackage[nolist]{acronym}

\usepackage{hyperref}

\usepackage[preprint]{icml2024}

\usepackage{amsmath}
\usepackage{amssymb}
\usepackage{mathtools}
\usepackage{amsthm}

\usepackage[capitalize,noabbrev]{cleveref}

\theoremstyle{plain}

\theoremstyle{definition}

\theoremstyle{remark}

\usepackage[textsize=tiny]{todonotes}

\usepackage{url}
\usepackage{xcolor} 
\definecolor{urlcolor}{RGB}{255,20,147} 
\hypersetup{urlcolor=urlcolor}

\icmltitlerunning{Large-Scale Dataset Pruning in Adversarial Training}

\begin{document}
    
    \twocolumn[
    \icmltitle{Large-Scale Dataset Pruning in Adversarial Training through Data Importance Extrapolation}
    
    
    
    \icmlsetsymbol{equal}{*}
    
    \begin{icmlauthorlist}
    \icmlauthor{Björn Nieth}{fau}
    \icmlauthor{Thomas Altstidl}{fau}
    \icmlauthor{Leo Schwinn}{equal,tum}
    \icmlauthor{Björn Eskofier}{equal,fau}
    \end{icmlauthorlist}
    
    \icmlaffiliation{fau}{Friedrich-Alexander University Erlangen-Nürnberg}
    \icmlaffiliation{tum}{Technical University Munich}
    
    \icmlcorrespondingauthor{Björn Nieth}{Bjoern.Nieth@fau.de}
    
    \icmlkeywords{Adversarial Training, Data Pruning}
    \centering
    
    \textcolor{red}{\url{https://github.com/BjoernNieth/LS-Dataset-pruning-in-AT}}
    \vskip 0.3in
    ]
    
    
    
    \printAffiliationsAndNotice{\icmlEqualContribution} 

    
    \begin{abstract}
    Their vulnerability to small, imperceptible attacks limits the adoption of deep learning models to real-world systems. Adversarial training has proven to be one of the most promising strategies against these attacks, at the expense of a substantial increase in training time. With the ongoing trend of integrating large-scale synthetic data this is only expected to increase even further. Thus, the need for data-centric approaches that reduce the number of training samples while maintaining accuracy and robustness arises. While data pruning and active learning are prominent research topics in deep learning, they are as of now largely unexplored in the adversarial training literature.
    We address this gap and propose a new data pruning strategy based on extrapolating data importance scores from a small set of data to a larger set. In an empirical evaluation, we demonstrate that extrapolation-based pruning can efficiently reduce dataset size while maintaining robustness.
    \end{abstract}
    \section{Introduction}
    
    Modern deep learning models show impressive capabilities in many different application areas like image classification \cite{he_deep_2015}, time-series processing~\cite{nguyen2021time, dumbach2023artificial} or natural language processing \cite{radfordImprovingLanguageUnderstanding}. A major limiting factor in deploying these models is their vulnerability to adversarial examples, maliciously crafted perturbations designed to compromise the predictions of a model~\cite{goodfellow_explaining_2015, schwinn2024soft}. 
    Different defenses were proposed to robustify networks against adversarial examples but the majority of these approaches were later shown to be ineffective~\cite{croce_reliable_2020, schwinn2023adversarial}. 
    
    One prevailing defense strategy is adversarial training (AT)~\cite{goodfellow_explaining_2015, madry_towards_2019, xhonneux2024efficient}. AT is a training paradigm where adversarial examples are directly incorporated into a model's training. 
    Multiple approaches have been employed to increase the performance of adversarial training, like more sophisticated loss functions \cite{zhang_theoretically_2019, wang_improving_2020}, increasing model capacity \cite{madry_towards_2019} or using \acf{SWA} \cite{izmailov_averaging_2019} during training \cite{gowal_improving_2021}. 
    Recent studies have demonstrated that using large quantities of synthetic data can lead to major robustness improvements \cite{rebuffi_fixing_2021, wang_better_2023, altstidl2023raising}.
    
    Nevertheless, compared to standard training, adversarial training considerably increases the training time \cite{madry_towards_2019}, which is further amplified by using synthetic data~\cite{rebuffi_fixing_2021}. 
    As a result, the computational costs associated with adversarial training will continue to increase, which provides a challenge for academic research.

    We argue that this problem can be tackled from a data-centric perspective. 
    Previous work illustrates the effectiveness of data selection algorithms, such as active learning or data pruning~\cite{coleman_selection_2020, yangDatasetPruningReducing2023}. Yet, these techniques are largely unexplored in adversarial machine learning, and the few existing works do not scale to large datasets~\cite{kaufmann_efficient_2022, li_less_2023, dolatabadi_adversarial_2023}.

    In this work, we address this problem and propose to extrapolate known data importance scores $u_{s_i}$ from a small set of training examples $\mathcal{S}$ to a larger set of previously unseen training data $\mathcal{D}$. To this end, we compute the feature embedding of $\mathcal{S}$ and $\mathcal{D}$ and calculate data importance scores for unseen samples $x_{d_i}$ as the average of their $k$-nearest neighbors in $\mathcal{S}$. As the computational complexity of the extrapolation is independent of the calculation of the data importance score, it allows us to employ accurate but expensive methods to calculate $u_{s_i}$. We demonstrate the efficiency of this approach using the recently proposed \acf{DU} pruning method in the setting of adversarial training. Our contributions are as follows:
    \begin{itemize}[noitemsep,topsep=0pt]
        \item[-] We adopt the data pruning strategy \ac{DU} to the domain of adversarial training and highlight differences in data importance between adversarial and standard training.
        \item[-] We propose a scalable method for data pruning by extrapolating a pruning metric from already existing scores to new samples using feature embeddings and $k$-nearest neighbor.
        \item[-] An empirical evaluation shows that our extrapolation-based pruning approach scales effectively to adversarial training with large synthetic datasets, achieving greater robustness than other methods with the same amount of data.
    \end{itemize} 
    
\subsection{Background}

Let \(f: \mathbb{R}^d \mapsto \mathbb{R}^C\) denote a differentiable classifier implemented by a neural network, where \(C \in \mathbb{N}\) is the number of classes. The prediction probability of a class \(y \in C\) given input \(x \in \mathbb{R}^d\) for \(f\) is denoted as \(p_f(y|x,\theta)\), where \(\theta\) are the parameters of \(f\). During training the parameters are continuously updated for $K$ epochs. We refer to $\theta_k$ as the parameters at epoch $k$. When trained adversarially, the input \(x \in \mathbb{R}^d\) is pertubed to \(\Tilde{x}_{\theta_k} \in \mathbb{R}^d\) using an adversarial attack $\mathcal{A}_{\epsilon,p}: \mathbb{R}^d \mapsto \mathbb{R}^d$, which aims to misclassify the input while keeping $\lVert x - \Tilde{x}_{\theta_i} \rVert_p < \epsilon$. The maximum magnitude of the perturbation $\epsilon$ and the norm $\lVert \cdot \rVert_p$ are design parameters of the attack chosen in advance.
    
In this study, we extend the heuristic data selection method \acf{DU}, initially described by~\citet{he_large-scale_2023}, to adversarial training with large datasets. The \ac{DU} metric is chosen due to its efficiency and minimal impact on model performance. \ac{DU} quantifies the fluctuating overall uncertainty $u \coloneq U(x,y)$ of each sample $x$ with class label $y$ throughout the training process. The authors define the prediction uncertainty at epoch $k = [J, \ldots, K]$ over a window of $J$ epochs as 
\begin{align}
    P_k(x,y) &= \sqrt{\frac{\sum^{J-1}_{j=0}[p_f(y|x,\theta_{k-j})-\mu_k(x,y)]^2}{J-1}},\\ 
    \mu_k(x,y) &= \frac{\sum^{J-1}_{j=0}p_f(y|x,\theta_{k-j})}{J}.
    \label{eq:Prediction_Uncertainty}
\end{align}

Then the overall \ac{DU} is defined as

\begin{equation}
    U(x,y) = \frac{\sum^{K}_{k=J}P_k(x,y)}{K-J}.
    \label{eq:Dynamic_Uncertainty}
\end{equation}
Intuitively, we can interpret the \ac{DU} as the average of the prediction uncertainty over a sliding window of size \(J\) for \(K\) epochs, where the prediction uncertainty is the standard deviation of the confidence in each window.

To adapt \ac{DU} to adversarial training, we simply exchange $x$ with $\Tilde{x}_{\theta_{k-j}}$ for every training sample. Namely, for every input, we first obtain its adversarial perturbed versions and subsequently use them to calculate the prediction uncertainty. In the remainder of this work, we denote the DU score from a standard training model as $U$. For an adversarial trained model, we refer to the \ac{DU} score from clean predictions as $U_{adv}$ and from adversarial predictions as $\Tilde{U}_{adv}$, the latter of which uses the perturbed $\Tilde{x}_{\theta_{k-j}}$ to compute the score.


\section{Extrapolating data importance scores}

%
    
In our work, we want to employ \ac{DU} pruning on large-scale synthetic datasets consisting of millions of images. Unfortunately, the calculation of the \ac{DU} for each sample $x$ requires certainties for each training epoch to be known and hence a network would need to be trained on the whole dataset. This makes it unsuitable for pruning, where the intent is to reduce the number of training images seen by the network during training. We thus devise an extrapolation procedure.

Let $x_{s_i} \in \mathcal{S}$ be the source set for which we compute the \ac{DU} scores, and $x_{d_i} \in \mathcal{D}$ be the destination set with unknown \ac{DU} scores with $\mathcal{S} \cap \mathcal{D} = \emptyset$. Then we are interested in extrapolating the \ac{DU} scores $\{u_{d_i} | i \in 1, \ldots, |\mathcal{D}|\}$ from the known $\{u_{s_i} | i \in 1, \ldots, |\mathcal{S}|\}$.

We base our extrapolation on a $k$-nearest neighbor search using an embedding function $\mathcal{E}: \mathbb{R}^d \mapsto \mathbb{R}^{d'}$, which embeds the input into a lower-dimensional space $d' \ll d$. Given the indices $j$ of the $k$ closest embeddings $e_{s_j}$ we then compute the extrapolated score as $u_{d_i} = \frac{1}{k} \sum_j u_{s_j}$. One could use the classifier originally trained on the source set for a straightforward encoding approach to derive the \ac{DU} scores for extrapolation. We note that this approach could easily be extended to more complex embedding functions, such as foundation models. For more details, see Section \ref{sec:experiment}.
    
\section{Experiment Setup}\label{sec:experiment}

In the following, we present hyperparameter settings for the conducted experiments.

\textbf{Dataset.} For all of our experiments, we use the CIFAR-10 \cite{krizhevsky_learning_nodate} dataset, which is commonly used for adversarial training and evaluation~\cite{madry_towards_2019, wongFastBetterFree2020,schwinn2021dynamically, wang_better_2023}. For the synthetic dataset, we use the 5 million synthetic CIFAR10 images provided by \citet{wang_better_2023} and subsample them to $2$ million samples while keeping the original class balance. Moreover, we sample a subset of $100,000$ images from the synthetic dataset as the extrapolation source set $\mathcal{S}$. To compare our extrapolation approach to the calculation of ground truth \ac{DU}, we sample a second, mutually exclusive set of $50,000$ images $\mathcal{D}_{test}$.    

\textbf{Model and training.} Our architecture and training procedures are the same as in~\cite{wang_better_2023}. We use a Wide-ResNet-28-10 for all experiments \cite{zagoruyko_wide_2017}. The model is trained using the TRADES \cite{zhang_theoretically_2019} framework with $\beta = 5$ and label smoothing of $0.1$. For the optimizer, we use Stochastic Gradient Descent with a Nesterov momentum of $0.9$ and a weight decay of $0.0005$. We use the super-convergence scheduler from \citet{smith_super-convergence_2018} with a maximum learning rate of $0.2$. On the model parameters, we apply \ac{SWA} implemented as a decay with a decay rate of $0.995$ and a warm-up phase of $2.5\%$ of the total update steps. We use common augmentations consisting of random crops and horizontal flips \cite{he_deep_2015}. 

\textbf{Adversarial training.} For the generation of the adversarial examples, we use 10 attack iterations in every training step. As in prior work, we conduct experiments using $\ell_{\infty}$-norm untargeted attacks with $\epsilon = 8/255$ and $\ell_{2}$-norm attacks with \(\epsilon = 128/255\) and choose a step size of $2/255$ and $32/255$, respectively.

We deploy a hold-out validation set to select the optimal epoch for our model parameters. To be comparable to \citet{wang_better_2023}, we also select the first 1024 images of the original CIFAR-10 dataset for our validation dataset. We evaluate the robust accuracy of our model using Auto Attack on the test set of CIFAR-10~\cite{croce_reliable_2020}.

\textbf{Dynamic uncertainty and data importance extrapolation.} In preliminary experiments on standard CIFAR-10 training, we compared the effectiveness of pruning using forgetting scores~\cite{toneva2018empirical}, Error L2-Norm
(EL2N) scores \cite{paul2021deep}, and Dynamic Uncertainty (DU) pruning~\cite{he_large-scale_2023}. \ac{DU} performed best in our experiments for low ($25\%$) and high ($50\%$) pruning regimes, and thus, we decided to use it for the remaining experiments. 

For the calculation of the \ac{DU} scores, we use the proposed window size of \(J=10\) from \cite{he_large-scale_2023}. 

To extrapolate these data importance scores, we use the following setup. 
As possible encoding strategies, we deploy the multipurpose visual feature model DINOv2 \cite{oquab_dinov2_2023}, the contrastive learning deduplication model SSCD \cite{pizzi_self-supervised_2022}, and a ResNet backbone of a CIFAR-10 classifier. All of these models learn an embedding of semantically meaningful features. For the distance metric of the KNN algorithm, we either use cosine similarity or Euclidian distance. The last hyperparameter is the number of neighbors $k$. In a grid search, we determine the best combination of the hyperparameters for extrapolating data importance scores.

\begin{figure}
     \centering
     \includegraphics{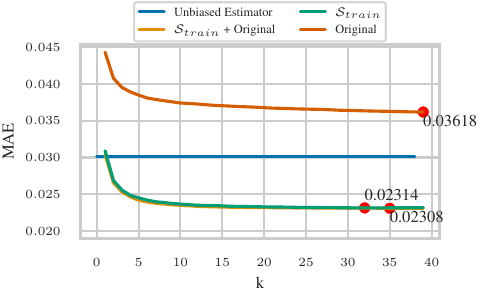}
     \caption{Mean Absolute Error (MAE) for different data importance extrapolation settings. $k$ denotes the number of nearest neighbors used for extrapolation.}
     \label{fig:Parameter_k}
 \end{figure}

To compute the evaluation metric, we use the subsampled $100,000$ synthetic images from $\mathcal{S}_{train}$ and train a classifier for each threat model to obtain the ground truth \ac{DU} scores of $\mathcal{S}_{train}$. We further take into account the \ac{DU} scores obtained from a separate training run on the original images.
For the aforementioned grid search, we split $10,000$ samples from $\mathcal{S}_{train}$ and used them as the validation set that we extrapolate to. Figure~\ref{fig:Parameter_k}  illustrates the mean absolute error (MAE) for various selections of the source dataset $\mathcal{S}$ and different numbers of neighbors $k$. The results indicate that using only data from the original CIFAR-10 dataset yields poorer extrapolation outcomes compared to using the larger synthetic dataset $\mathcal{S}_{train}$. The extrapolation can be further improved by merging scores from both the original and synthetic datasets. Results are compared to the unbiased estimator, for which we use the mean \ac{DU} score of the training samples of $\mathcal{S}_{train}$. 


Exact hyperparameter settings are given in Appendix~\ref{tbl:hyperparameter}.

\begin{figure}
    \centering
    \includegraphics{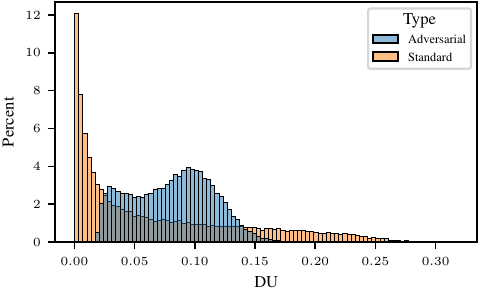}
    \caption{Distribution of adversarial \acf{DU} scores vs. standard \ac{DU} scores.}
    \label{fig:distribution_standard_vs_adversarial_du}
\end{figure}

\section{Results}

This paper aims to enable effective pruning in adversarial training with large datasets by extrapolating data importance scores from existing pruning methods. To this end, we first conduct several experiments to investigate the utility of the \ac{DU} metric in adversarial training without synthetic data. To evaluate if data importance extrapolation generalizes to other metrics, we further construct our own pruning metric and demonstrate its performance for robust training. Lastly, we evaluate our extrapolation approach on large datasets.  

\subsection{Dynamic uncertainty analysis}

\begin{table}
    \caption{Comparison of different pruning approaches on the original CIFAR-10 dataset for $\ell_{2}$ and $\ell_{\infty}$ adversarial training. Either $25\%$ or $50\%$ of the training data is pruned. Training on the full dataset (Baseline) is compared to different dynamic uncertainty scores $U$, $U_{adv}$, $\Tilde{U}_{adv}$, and FP. Random pruning is additionally provided as a baseline.}
    \label{tbl:L2_normal}
    \begin{center}
    \begin{small}
    \begin{sc}
    \resizebox{0.48\textwidth}{!}{
        \begin{tabular}{lllrrr}
        \toprule[\heavyrulewidth]
        Norm & Pruned & Experiment                       & Robust      & Clean & Speed    \\ \midrule[\lightrulewidth]
        \multirow{11}{*}{$\ell_2$} & 0\% & Baseline                         & 72.97\%     & 88.89\%  & 1.00           \\ \cmidrule(l){2-6}
        & \multirow{5}{*}{25\%} & ${U}$               & 71.50\%   & 87.42\%  & 1.33 \\
        & & ${U}_{adv}$ & 72.41\%   & 88.59\%  & 2.63 \\
        & & $\Tilde{U}_{adv}$           & 72.81\%     & 88.79\%  & 1.73 \\
        & & FP          &   \textbf{72.91}\% & 88.86\%  & 2.07  \\
        & & Random                       & 67.20\%  & 84.71\%  & 1.87 \\ \cmidrule(l){2-6}
        & \multirow{5}{*}{50\%} & ${U}$ & 62.93\%    & 82.94\%  & 2.48 \\
        & & ${U}_{adv}$                  & 63.07\%    & 85.47\%  & 2.85 \\
        & & $\Tilde{U}_{adv}$            & \textbf{66.93}\%    & 86.08\%  & 3.22 \\
        & & FP          &  66.22\%    &  87.19\% & 1.87  \\
        & & Random                       & 64.21\%    & 82.25\%  & 2.14 \\
        \midrule
        \multirow{11}{*}{$\ell_\infty$} & 0\% & Baseline       & 54.19\%     & 84.42\%  & 1.00           \\ \cmidrule(l){2-6}
        & \multirow{5}{*}{25\%} & ${U}$ & 50.22\%     & 81.72\%  & 1.55 \\
        & & ${U}_{adv}$ & 50.10\%     & 83.65\%  & 1.53 \\
        & & $\Tilde{U}_{adv}$    & \textbf{52.58}\%     & 83.90\%  & 1.55 \\
        & & Random      & 47.58\%     & 79.22\%  & 2.83 \\ \cmidrule(l){2-6}
        & \multirow{5}{*}{50\%} & ${U}$ & 39.47\%     & 76.21\%  & 2.27 \\
        & & ${U}_{adv}$ & 33.86\%     & 75.44\%  & 2.18 \\
        & & $\Tilde{U}_{adv}$    & 41.04\%     & 80.19\%  & 2.55 \\
        & & Random      & \textbf{44.93}\%     & 77.38\%  & 2.96 \\
        \bottomrule
        \end{tabular}
        }
    \end{sc}
    \end{small}
    \end{center}
    \vskip -0.1in
\end{table}

As a preliminary experiment, we investigate the \ac{DU} metric for adversarial training. Figure~\ref{fig:distribution_standard_vs_adversarial_du} highlights differences in the distributions of \ac{DU} scores between standard and adversarial training. For the standard \ac{DU} most of the scores are close to zero and the distribution has a relatively long tail compared to the adversarial distribution. The variations in the calculated \ac{DU} scores point towards differences in the predicted certainties between these two training approaches. 

\begin{figure*}
    \centering
    \includegraphics[width=\linewidth]{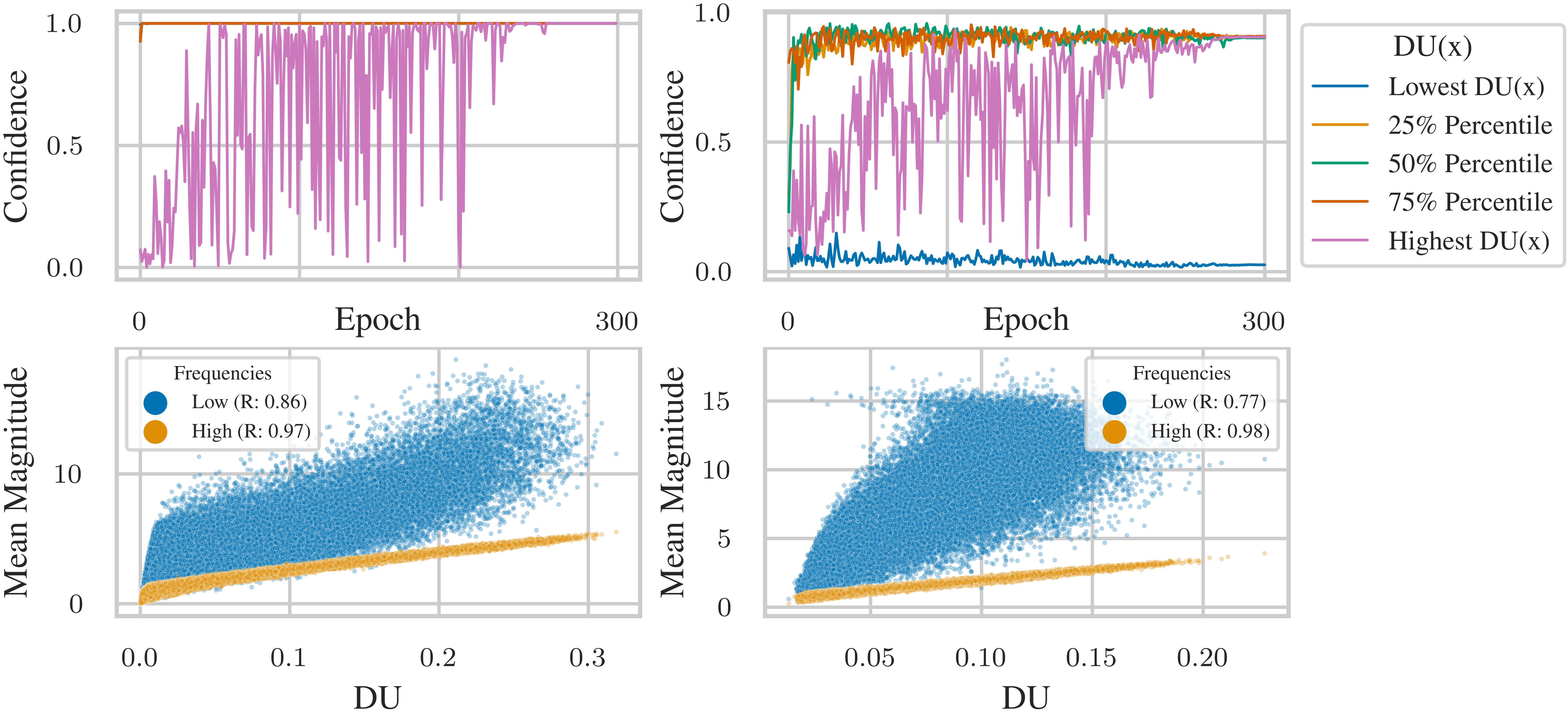}
    \subfloat[Standard training\label{fig:confidences_standard}]{\hspace{.38\linewidth}}
    \subfloat[$\ell_2$ adversarial training \label{fig:Confidences_adversarial}]{\hspace{.55\linewidth}}
    \caption{The upper two plots show predicted certainties for individual samples during standard training (left) and adversarial training (right). The images below visualize
    the relationship between the DU of a sample and the mean magnitude of the DFT calculated on the training dynamics of the sample. We distinguish the mean magnitude of the low-frequency spectrum (frequencies 1-10) and the high-frequency spectrum (11-150). $R$ denotes the Pearson correlation between the magnitude and the DU. }\label{fig:certainties_both}
\end{figure*}

To investigate this further, we compare how the predicted certainties of adversarially and standardly trained samples evolve during training. We analyze samples with a wide range of different \ac{DU} scores, including very low, average, and high ones. In the upper part of Figure~\ref{fig:certainties_both}, we compare the predicted certainties of different samples from standard and adversarial training. In standard training, the samples converge to a certainty close to one, and except for the samples with a high \ac{DU} score, the convergence is very smooth and does not show high oscillations. For adversarial training, there is generally a high level of oscillations present, which in turn will influence the \ac{DU}. 

The \ac{DU} computes the average over the standard deviation in a sliding windows approach. This can be interpreted as a high-pass filter on the training dynamic, where small window sizes will prevent long-term training trends (e.g., slow convergence) from influencing the \ac{DU} score. To analyze this further, we compute the \acf{DFT} of the training dynamics for standard training and $\ell_{2}$-based adversarial training. In Figure~\ref{fig:certainties_both}, we plot the sum of the magnitude of the lower $1-10$ (Low) and higher $11-150$ (High) frequencies of the DFT vs the \ac{DU} score. It can be seen that for standard training, high frequencies in the training dynamic are more indicative of a high \ac{DU} score (Pearson R=$0.97$) compared to low frequencies (R=$0.86$). This supports our perspective of \ac{DU} acting as a high-pass filter on the training dynamics. While this trend still holds for $\ell_{2}$-based adversarial training, lower frequencies are even less correlated with high \ac{DU} scores in this case. 

We propose a simple pruning metric called \acf{FP}, which considers both short-term trends and long-term training dynamics. In \ac{FP} we first calculate the DFT from the training dynamics and calculate the data importance scores as the magnitude of the DFT for the respective sample. By considering all frequencies in the training dynamics, the \ac{FP} captures both global convergence of the predicted certainties and local oscillations. The main purpose of the \ac{FP} metric is to investigate if other data importance scores beyond \ac{DU} can be extrapolated for data pruning. Due to the high computational requirements of adversarial training, we constrain \ac{FP} related experiments to $\ell_2$-based adversarial training. 

    
\subsection{Dynamic uncertainty in adversarial training}

\begin{table}
    \caption{Evaluation of extrapolated dynamic uncertainty scores on a $2$ million sample synthetic CIFAR-10 dataset. $50\%$ of the $2$ million synthetic samples are pruned for each method. BALANCED indicates that class balance was maintained during pruning. Results are compared to the same training setup with $1$ million images from~\citet{wang_better_2023} and to random pruning.}
    \label{tbl:extrapolation_results}
    \begin{center}
    \begin{small}
    \begin{sc}
    \resizebox{0.48\textwidth}{!}{
        \begin{tabular}{llll}
        \toprule[\heavyrulewidth]
        Norm                  & Experiment  & Robust                & Clean             \\ \midrule[\lightrulewidth]
        \multirow{6}{*}{\(\ell_2\)}   & 1m \cite{wang_better_2023}                                  & 79.98\%                & 93.76\%                \\ \cmidrule(l){2-4}
          & 1m Random                                          & 80.79\%   & 94.20\%  \\ \cmidrule(l){2-4}
          & 1m $\Tilde{U}_{adv}$                                                 & 80.89\%                               & 94.37\%                \\
          & 1m $\Tilde{U}_{adv}$ Balanced                                       & 81.28\%                               & 94.22\%                \\
          & 1m FP                                                & 81.00\%                               & 94.43\%                \\
          & 1m FP Balanced                                       & \textbf{81.38}\%                               & 94.47\%                \\  \midrule[\lightrulewidth]
        \multirow{6}{*}{\(\ell_\infty\)} & 1m \cite{wang_better_2023}                                  & 63.35\%                     & 91.12\%                \\ \cmidrule(l){2-4}
          & 1m Random & 63.13\%   & 90.86\% \\ \cmidrule(l){2-4}
          & 1m $\Tilde{U}_{adv}$                                               & 62.29\%                               & 91.22\%                \\
          & 1m $\Tilde{U}_{adv}$ Balanced                                       & \textbf{63.56}\%                               & 90.50\%                \\
        \bottomrule
        \end{tabular}
        }
    \end{sc}
    \end{small}
    \end{center}
    \vskip -0.3in
\end{table}

Before scaling to large datasets, we investigate the effectiveness of \ac{DU} on a smaller scale. Thus, we adversarially train a model with the configurations described in Section~\ref{sec:experiment} for $300$ epochs on the original CIFAR-10 dataset. This experiment is conducted for both $\ell_2$ and $\ell_\infty$ threat models. From the training of this model, we store both the clean \ac{DU} scores $U_{adv}$ and adversarial \ac{DU} scores $\Tilde{U}_{adv}$ for each sample and epoch $k$. As a reference, we also train a model using standard training and save the clean scores $U$.

After obtaining the necessary \ac{DU} scores, we adversarially train a network using the pruned dataset. The resulting performances for both threat models are shown in Table~\ref{tbl:L2_normal}. Here, ${U}_{adv}$ refers to the data importance scores based on the clean certainties obtained from the adversarially trained model, $\Tilde{U}_{adv}$ refers to the data importance scores based on the adversarial certainties obtained from the adversarially trained model, and $U$ refers to the standard training regime. Additionally, \ac{FP} refers to the proposed frequency pruning method. As an additional baseline, we include values for a randomly subsampled dataset.

We observe that, \ac{DU} pruning and \ac{FP} achieve the highest robustness for the adversarial certainties $\Tilde{U}_{adv}$ obtained from adversarial training and hence these will be used in subsequent experiments. When pruning with the \ac{DU} metric, we can achieve training speed-ups of $1.73$ and $1.55$ for $\ell_2$ and $\ell_\infty$, respectively, with only minor degradations in robustness. Using the \ac{FP} metric we can achieve a speed-up of $2.07$ for $\ell_2$ with only a minor degradation in robustness. 

\begin{figure}
    \centering
    \includegraphics{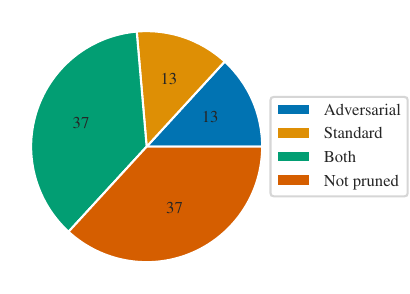}
    \caption{Overlap between the sets pruned by $\Tilde{U}_{adv}$ vs. $U$ for 50\% pruning on CIFAR-10. Note that only for two thirds of all images they result in the same pruning decision.}
    \label{fig:adversarial_DU_vs_standard_DU}
\end{figure}

Moreover, we explored the overlap between the samples pruned using the $\Tilde{U}_{adv}$ scores and the standard dynamic uncertainty scores $U$. Both sets contain the most important samples for adversarial and standard training, respectively. For $50\%$ pruning, this is visualized in Figure~\ref{fig:adversarial_DU_vs_standard_DU}. The different \ac{DU} scores lead to substantially different pruning behaviors, with only $37\%$ overlap between the two sets. This result indicates considerable differences concerning data importance between the standard and adversarial training settings.

\subsection{Extrapolation experiments}\label{sec:res-extrapolation}

For our extrapolation approach on large-scale synthetic datasets, we use the optimal hyperparameters from the grid search to extrapolate the data importance scores to a synthetic dataset of 2 million images. Using our extrapolated data importance scores and random subsampling, we train a model using 1 million synthetic images out of 2 million. For the extrapolated pruning approaches, we also run an experiment, where the data importance scores are only compared inside of each class, denoted as class-balanced pruning. Results are compared against the model from \citet{wang_better_2023} using a similar amount of synthetic data.\\
From previous work it is known that adversarial training using synthetic data sets does not exhibit robust overfitting \cite{rebuffi_fixing_2021,wang_better_2023}. We compare the robust accuracy achieved by the model for a fixed number of training steps to evaluate the effectiveness of our approach.  

We use our extrapolation strategy to estimate the data importance scores for the large-scale synthetic dataset and run the experiments for both threat models. The results are shown in Table~\ref{tbl:extrapolation_results}. For both threat models, our approach beats the baseline from \citet{wang_better_2023} and the random baseline. Under both threat models, the best pruning approach uses class-balanced pruning. For the \(\ell_2\)-norm threat model, our introduced \ac{FP} metric yields the best performing model, having a higher clean and robust accuracy than any other of the trained models.  

\subsection{Limitations}

In the following, we describe relevant limitations of our data pruning approach with respect to how scores are extrapolated and the chosen pruning metric.
    
\begin{figure}
    \centering
    \includegraphics[width=0.5\textwidth]{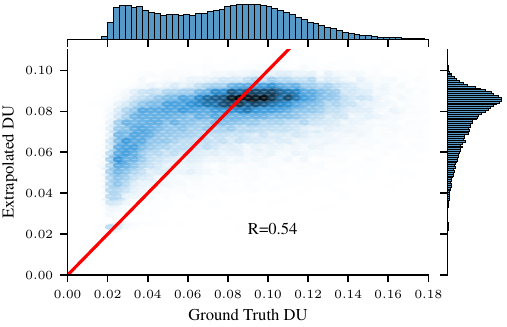}
    \caption{Extrapolated \acf{DU} score distribution compared to ground truth \ac{DU} score distribution. Extrapolated scores are biased toward the mean of the ground truth distribution.}
    \vspace{-1em}
    \label{fig:extrapolation_scores_vs_GT_scores}
\end{figure}

\paragraph{Extrapolation model} Here, we analyze the performance gap between Ground Truth (GT) \ac{DU} and our extrapolated \ac{DU} scores. We use $50,000$ independently sampled synthetic images for this experiment $\mathcal{D}_{test}$ (see Section~\ref{sec:experiment}).
Figure~\ref{fig:extrapolation_scores_vs_GT_scores} compares the GT scores from the independent test set with the extrapolated scores. The red line indicates the line of optimal prediction, where the predicted scores exactly match the GT scores. It is evident that the $k$-nearest neighbor extrapolation model utilized in our experiments is overly simplistic, failing to accurately predict the data importance scores for the samples. Table~\ref{tbl:extrapolated_vs_GT} presents our performance analysis of the data importance scores on the test set. At lower pruning percentages, the extrapolated scores are comparable to the Ground Truth (GT) scores; however, performance deteriorates at higher percentages. Note that calculating the GT scores requires training on the complete dataset, making this approach infeasible in practice.
A more sophisticated extrapolation model would likely yield a more accurate estimate of these scores, enhancing the performance of the downstream models. 

\begin{table}
    \caption{Comparisson between the utility of extrapolated vs. ground truth dynamic uncertainty values on a dataset of 50 thousand synthetic images $\mathcal{D}_{test}$. Specifically, when large amounts of data are pruned, ground truth values are superior to extrapolation.}
    \label{tbl:extrapolated_vs_GT}
    \begin{center}
    \begin{small}
    \begin{sc}
        \begin{tabular}{llll}
                    \toprule
        Experiment                              & Robust  & Clean \\ \midrule
        Baseline Original                                & 54.19\%         & 84.42\%    \\
        Baseline $\mathcal{S}_{train}$                         & 56.34\%        & 86.27\%    \\
        Baseline $\mathcal{D}_{test}$                          & 53.32\%        & 83.55\%    \\ \midrule
        25\%DU-AT-AT               & 52.26\%         & 81.95\%    \\
        25\%DU-AT-AT-Extra. & 50.39\%         & 80.86\%    \\
        25\% Random                             & 47.95\%         & 80.43\%    \\ \midrule
        50\%DU-AT-AT               & 48.52\%        & 77.67\%    \\
        50\%DU-AT-AT-Extra. & 42.11\%        & 67.10\%    \\
        50\% Random                             & 45.37\%         & 78.19\%    \\ \bottomrule     
        
        \end{tabular}
    \end{sc}
    \end{small}
    \end{center}
    \vskip -0.1in
\end{table}

\paragraph{Pruning metric} In our experiments, we primarily used the \ac{DU} metric to assess data importance. Additionally, we introduced a new metric, \ac{FP}, based on the frequency spectrum of training dynamics. Despite its simplicity, \ac{FP} performed remarkably well. While the performance of models trained with either data importance metric is comparable, our experiments show that they consider different samples for pruning, with little overlap between the pruned sets. This suggests that further research into heuristic pruning approaches is necessary, as the attribution of each data sample to the model performance remains largely unknown. 

A promising approach may include extrapolating data attribution metrics, which provide deeper insights into data importance but demand significant computational resources~\cite{ilyas2022datamodels}. Extrapolation could potentially lower these computational demands while preserving the advantages of data attribution.


\paragraph{Performance Evaluation} Due to the considerable cost of adversarial training, we compared our approach using a relatively small model and large synthetic dataset. Scaling to larger datasets, such as ImageNet, was out of the scope of this work and we leave it to future research.

\section{Related Work}

\paragraph{Adversarial Training} In the seminal work of \citet{szegedy_intriguing_2014}, the authors discovered the vulnerability of deep learning models against adversarial examples. 
Adversarial training (AT) was one of the first defenses proposed against these attacks~\cite{goodfellow_explaining_2015} and further enhanced in \citet{madry_towards_2019}. 
Later work proposed more sophisticated loss functions for adversarial training, like TRADES \cite{zhang_theoretically_2019} or MARTS \cite{wang_improving_2020}. Other techniques like increasing model capacity \cite{madry_towards_2019} or \ac{SWA} \cite{gowal_improving_2021} are employed to increase the robustness of a model. Data augmentation is a debated topic in adversarial training, with different works pointing in different directions \cite{carmon_unlabeled_2022, rice_overfitting_2020, gowal_uncovering_2021, rebuffi_fixing_2021,li_data_2023, wang_better_2023}.\\
\citet{schmidt_adversarially_2018} illustrate, that adversarial generalization needs more data than standard training from a theoretical viewpoint. In line with this observation, later work demonstrated that the adversarial robustness can be increased through additional unlabeled data \cite{carmon_unlabeled_2022, gowal_uncovering_2021}. In recent work, generative models were employed, to produce additional synthetic samples, which has resulted in state-of-the-art performance for robust models \cite{rebuffi_fixing_2021, wang_better_2023, peng_robust_2023}. Yet, this trend has resulted in considerably increased training times. 

\paragraph{Data Selection in Adversarial Training} The work on data selection in adversarial training is sparse. 
In general, three different categories of data selection in robust training are available. The first strategy involves dynamically deciding during training for which samples, the adversarial perturbations should be applied and which samples are used without perturbation~\cite{hua_bullettrain_2021,chen_data_2024}. The second line of work includes corset selection strategies applied to adversarial training~\cite{killamsetty_grad-match_2021, dolatabadi_adversarial_2023}. The last line of work consists of heuristic data pruning in adversarial training~\cite{kaufmann_efficient_2022, li_less_2023}.\\
For the first category, the most influential work is BulletTrain \cite{hua_bullettrain_2021}. In BulletTrain, only a full perturbation is calculated for samples close to the decision boundary of the model. This selection is done dynamically during the training and can considerably speed up the training without losing performance. In \citet{chen_data_2024} the authors calculate a few iterations of the adversarial perturbation and only continue the calculation if the loss for the sample is increasing. In \citet{gupta_we_2023} the authors only calculate the adversarial examples for samples that are not robust. These samples are selected through an efficient but weak attack on the training data. However, these works do not directly address the problem of increasingly large datasets. Moreover, \citet{gupta_we_2023} did not use a modern adversarial training setup or a standardized test for adversarial performance evaluation. Therefore the impact of their work is hard to estimate.

The only work surrounding adversarial core set selection evolves around the adaptation of the GRAD-MATCH \cite{killamsetty_grad-match_2021} algorithm to the domain of adversarial training. Both \citet{dolatabadi_adversarial_2023} and \citet{li_less_2023} give an adversarial adaptation of GRAD-MATCH. Both approaches speed up the training time, but also substantially decrease the resulting model's performance. In \citet{xu_efficient_2023}, the authors deploy a core set selection algorithm to the related field of adversarial contrastive learning. Still, due to the high computational complexity of coreset selection, these approaches do not efficiently scale to large datasets.

Closer to this work, Adv-GLISTER \cite{li_less_2023} adapts the GLISTER \cite{killamsetty_glister_2021} algorithm for adversarial training. Adv-GLISTER speeds up the training time but considerably reduces the performance of the model. The most comparable work to this paper is \citet{kaufmann_efficient_2022}. Here, the authors test different data-pruning strategies for robust training. After a fixed number of epochs, a certain amount of data is dropped according to a data selection strategy. However, while this approach decreases the training time, it also decreases accuracy and robustness.
    
\section{Conclusion}

We propose a novel method to scale data pruning for large datasets, enabling adversarial training with extensive synthetic data. Our approach shows that data importance can be extrapolated from small to large datasets, circumventing the need for expensive pruning methods on the entire dataset. We demonstrate the effectiveness of applying our extrapolation method to \ac{DU} in adversarial training contexts. Additionally, we compare samples pruned for standard and adversarial training, revealing significant differences in data importance. Our experiments underscore the potential of extrapolating computationally intensive scores to unseen data, paving the way for future exploration of more complex metrics like data attribution scores.

\section{Acknowledgement}

We thank Gauthier Gidel and David Dobre for their assistance and feedback. Leo Schwinn gratefully acknowledges funding by the Deutsche Forschungsgemeinschaft (DFG, German Research Foundation) - Projectnumber 544579844.


    \bibliography{example_paper}

\begin{thebibliography}{46}
\providecommand{\natexlab}[1]{#1}
\providecommand{\url}[1]{\texttt{#1}}
\expandafter\ifx\csname urlstyle\endcsname\relax
  \providecommand{\doi}[1]{doi: #1}\else
  \providecommand{\doi}{doi: \begingroup \urlstyle{rm}\Url}\fi

\bibitem[Altstidl et~al.(2023)Altstidl, Dobre, Eskofier, Gidel, and Schwinn]{altstidl2023raising}
Altstidl, T., Dobre, D., Eskofier, B., Gidel, G., and Schwinn, L.
\newblock Raising the {{Bar}} for {{Certified Adversarial Robustness}} with {{Diffusion Models}}.
\newblock \emph{arXiv preprint arXiv:2305.10388}, 2023.

\bibitem[Carmon et~al.(2022)Carmon, Raghunathan, Schmidt, Liang, and Duchi]{carmon_unlabeled_2022}
Carmon, Y., Raghunathan, A., Schmidt, L., Liang, P., and Duchi, J.~C.
\newblock Unlabeled data improves adversarial robustness.
\newblock \emph{NeurIPS 2019}, 2022.

\bibitem[Chen \& Lee(2024)Chen and Lee]{chen_data_2024}
Chen, E.-C. and Lee, C.-R.
\newblock Data filtering for efficient adversarial training.
\newblock \emph{Pattern Recognition}, 2024.

\bibitem[Coleman et~al.(2020)Coleman, Yeh, Mussmann, Mirzasoleiman, Bailis, Liang, Leskovec, and Zaharia]{coleman_selection_2020}
Coleman, C., Yeh, C., Mussmann, S., Mirzasoleiman, B., Bailis, P., Liang, P., Leskovec, J., and Zaharia, M.
\newblock Selection via proxy: {{Efficient}} data selection for deep learning.
\newblock \emph{ICLR 2020}, 2020.

\bibitem[Croce \& Hein(2020)Croce and Hein]{croce_reliable_2020}
Croce, F. and Hein, M.
\newblock Reliable evaluation of adversarial robustness with an ensemble of diverse parameter-free attacks.
\newblock \emph{arXiv preprint arXiv:2003.01690}, 2020.

\bibitem[Dolatabadi et~al.(2023)Dolatabadi, Erfani, and Leckie]{dolatabadi_adversarial_2023}
Dolatabadi, H.~M., Erfani, S., and Leckie, C.
\newblock Adversarial coreset selection for efficient robust training.
\newblock \emph{International Journal of Computer Vision}, 2023.

\bibitem[Dumbach et~al.(2023)Dumbach, Schwinn, L{\"o}hr, Elsberger, and Eskofier]{dumbach2023artificial}
Dumbach, P., Schwinn, L., L{\"o}hr, T., Elsberger, T., and Eskofier, B.~M.
\newblock Artificial intelligence trend analysis in german business and politics: a web mining approach.
\newblock \emph{International Journal of Data Science and Analytics}, pp.\  1--22, 2023.

\bibitem[Goodfellow et~al.(2015)Goodfellow, Shlens, and Szegedy]{goodfellow_explaining_2015}
Goodfellow, I.~J., Shlens, J., and Szegedy, C.
\newblock Explaining and harnessing adversarial examples.
\newblock \emph{ICLR 2015}, 2015.

\bibitem[Gowal et~al.(2021{\natexlab{a}})Gowal, Qin, Uesato, Mann, and Kohli]{gowal_uncovering_2021}
Gowal, S., Qin, C., Uesato, J., Mann, T., and Kohli, P.
\newblock Uncovering the limits of adversarial training against norm-bounded adversarial examples.
\newblock \emph{arXiv preprint arXiv:2010.03593}, 2021{\natexlab{a}}.

\bibitem[Gowal et~al.(2021{\natexlab{b}})Gowal, Rebuffi, Wiles, Stimberg, Calian, and Mann]{gowal_improving_2021}
Gowal, S., Rebuffi, S.-A., Wiles, O., Stimberg, F., Calian, D.~A., and Mann, T.
\newblock Improving robustness using generated data.
\newblock \emph{NeurIPS 2021}, 2021{\natexlab{b}}.

\bibitem[Gupta \& Narayan(2023)Gupta and Narayan]{gupta_we_2023}
Gupta, V. and Narayan, A.
\newblock Do we need entire training data for adversarial training?
\newblock \emph{arXiv preprint arXiv:2303.06241}, 2023.

\bibitem[He et~al.(2015)He, Zhang, Ren, and Sun]{he_deep_2015}
He, K., Zhang, X., Ren, S., and Sun, J.
\newblock Deep residual learning for image recognition.
\newblock \emph{arXiv preprint arXiv:1512.03385}, 2015.

\bibitem[He et~al.(2023)He, Yang, Huang, and Zhao]{he_large-scale_2023}
He, M., Yang, S., Huang, T., and Zhao, B.
\newblock Large-scale dataset pruning with dynamic uncertainty.
\newblock \emph{arXiv preprint arXiv:2306.05175}, 2023.

\bibitem[Hua et~al.(2021)Hua, Zhang, Guo, Zhang, and Suh]{hua_bullettrain_2021}
Hua, W., Zhang, Y., Guo, C., Zhang, Z., and Suh, G.~E.
\newblock {{BulletTrain}}: {{Accelerating}} robust neural network training via boundary example mining.
\newblock \emph{NeurIPS 2021}, 2021.

\bibitem[Ilyas et~al.(2022)Ilyas, Park, Engstrom, Leclerc, and Madry]{ilyas2022datamodels}
Ilyas, A., Park, S.~M., Engstrom, L., Leclerc, G., and Madry, A.
\newblock Datamodels: Predicting predictions from training data.
\newblock \emph{arXiv preprint arXiv:2202.00622}, 2022.

\bibitem[Izmailov et~al.(2019)Izmailov, Podoprikhin, Garipov, Vetrov, and Wilson]{izmailov_averaging_2019}
Izmailov, P., Podoprikhin, D., Garipov, T., Vetrov, D., and Wilson, A.~G.
\newblock Averaging weights leads to wider optima and better generalization.
\newblock \emph{arXiv preprint arXiv:1803.05407}, 2019.

\bibitem[Kaufmann et~al.(2022)Kaufmann, Zhao, Shumailov, Mullins, and Papernot]{kaufmann_efficient_2022}
Kaufmann, M., Zhao, Y., Shumailov, I., Mullins, R., and Papernot, N.
\newblock Efficient adversarial training with data pruning.
\newblock \emph{arXiv preprint arXiv:2207.00694}, 2022.

\bibitem[Killamsetty et~al.(2021{\natexlab{a}})Killamsetty, Sivasubramanian, Ramakrishnan, De, and Iyer]{killamsetty_grad-match_2021}
Killamsetty, K., Sivasubramanian, D., Ramakrishnan, G., De, A., and Iyer, R.
\newblock {{GRAD-MATCH}}: {{Gradient}} matching based data subset selection for efficient deep model training.
\newblock \emph{PMLR}, 2021{\natexlab{a}}.

\bibitem[Killamsetty et~al.(2021{\natexlab{b}})Killamsetty, Sivasubramanian, Ramakrishnan, and Iyer]{killamsetty_glister_2021}
Killamsetty, K., Sivasubramanian, D., Ramakrishnan, G., and Iyer, R.
\newblock {{GLISTER}}: {{Generalization}} based data subset selection for efficient and robust learning.
\newblock \emph{arXiv preprint arXiv:2012.10630}, 2021{\natexlab{b}}.

\bibitem[Krizhevsky(2009)]{krizhevsky_learning_nodate}
Krizhevsky, A.
\newblock Learning multiple layers of features from tiny images.
\newblock \emph{University of Toronto}, 2009.

\bibitem[Li \& Spratling(2023)Li and Spratling]{li_data_2023}
Li, L. and Spratling, M.
\newblock Data augmentation alone can improve adversarial training.
\newblock \emph{ICLR 2023}, 2023.

\bibitem[Li et~al.(2023)Li, Zhao, Lin, Kailkhura, and Goldhahn]{li_less_2023}
Li, Y., Zhao, P., Lin, X., Kailkhura, B., and Goldhahn, R.
\newblock Less is more: {{Data}} pruning for faster adversarial training.
\newblock \emph{CEUR WS 2023}, 2023.

\bibitem[Madry et~al.(2019)Madry, Makelov, Schmidt, Tsipras, and Vladu]{madry_towards_2019}
Madry, A., Makelov, A., Schmidt, L., Tsipras, D., and Vladu, A.
\newblock Towards deep learning models resistant to adversarial attacks.
\newblock \emph{ICLR 2018}, 2019.

\bibitem[Nguyen et~al.(2021)Nguyen, Chatterjee, Weinzierl, Schwinn, Matzner, and Eskofier]{nguyen2021time}
Nguyen, A., Chatterjee, S., Weinzierl, S., Schwinn, L., Matzner, M., and Eskofier, B.
\newblock Time matters: {{Time-aware}} lstms for predictive business process monitoring.
\newblock In Leemans, S. and Leopold, H. (eds.), \emph{Process Mining Workshops}, pp.\  112--123, Cham, 2021. Springer International Publishing.
\newblock ISBN 978-3-030-72693-5.

\bibitem[Oquab et~al.(2024)Oquab, Darcet, Moutakanni, Vo, Szafraniec, Khalidov, Fernandez, Haziza, Massa, {El-Nouby}, Assran, Ballas, Galuba, Howes, Huang, Li, Misra, Rabbat, Sharma, Synnaeve, Xu, Jegou, Mairal, Labatut, Joulin, and Bojanowski]{oquab_dinov2_2023}
Oquab, M., Darcet, T., Moutakanni, T., Vo, H., Szafraniec, M., Khalidov, V., Fernandez, P., Haziza, D., Massa, F., {El-Nouby}, A., Assran, M., Ballas, N., Galuba, W., Howes, R., Huang, P.-Y., Li, S.-W., Misra, I., Rabbat, M., Sharma, V., Synnaeve, G., Xu, H., Jegou, H., Mairal, J., Labatut, P., Joulin, A., and Bojanowski, P.
\newblock {{DINOv2}}: {{Learning}} robust visual features without supervision.
\newblock \emph{TMLR 2024}, 2024.

\bibitem[Paul et~al.(2021)Paul, Ganguli, and Dziugaite]{paul2021deep}
Paul, M., Ganguli, S., and Dziugaite, G.~K.
\newblock Deep learning on a data diet: Finding important examples early in training.
\newblock \emph{Advances in Neural Information Processing Systems}, 34:\penalty0 20596--20607, 2021.

\bibitem[Peng et~al.(2023)Peng, Xu, Cornelius, Hull, Li, Duggal, Phute, Martin, and Chau]{peng_robust_2023}
Peng, S., Xu, W., Cornelius, C., Hull, M., Li, K., Duggal, R., Phute, M., Martin, J., and Chau, D.~H.
\newblock Robust principles: {{Architectural}} design principles for adversarially robust {{CNNs}}.
\newblock \emph{arXiv preprint arXiv:2308.16258}, 2023.

\bibitem[Pizzi et~al.(2022)Pizzi, Roy, Ravindra, Goyal, and Douze]{pizzi_self-supervised_2022}
Pizzi, E., Roy, S.~D., Ravindra, S.~N., Goyal, P., and Douze, M.
\newblock A self-supervised descriptor for image copy detection.
\newblock \emph{CVPR 2022}, 2022.

\bibitem[Radford et~al.(2018)Radford, Narasimhan, Salimans, and Sutskever]{radfordImprovingLanguageUnderstanding}
Radford, A., Narasimhan, K., Salimans, T., and Sutskever, I.
\newblock Improving {{Language Understanding}} by {{Generative Pre-Training}}.
\newblock 2018.

\bibitem[Rebuffi et~al.(2021)Rebuffi, Gowal, Calian, Stimberg, Wiles, and Mann]{rebuffi_fixing_2021}
Rebuffi, S.-A., Gowal, S., Calian, D.~A., Stimberg, F., Wiles, O., and Mann, T.
\newblock Fixing data augmentation to improve adversarial robustness.
\newblock \emph{arXiv preprint arXiv:2103.01946}, 2021.

\bibitem[Rice et~al.(2020)Rice, Wong, and Kolter]{rice_overfitting_2020}
Rice, L., Wong, E., and Kolter, J.~Z.
\newblock Overfitting in adversarially robust deep learning.
\newblock \emph{PMLR 2020}, 2020.

\bibitem[Schmidt et~al.(2018)Schmidt, Santurkar, Tsipras, Talwar, and M{a}dry]{schmidt_adversarially_2018}
Schmidt, L., Santurkar, S., Tsipras, D., Talwar, K., and M{a}dry, A.
\newblock Adversarially robust generalization requires more data.
\newblock \emph{NeurIPS2028}, 2018.

\bibitem[Schwinn et~al.(2021)Schwinn, Nguyen, Raab, Zanca, Eskofier, Tenbrinck, and Burger]{schwinn2021dynamically}
Schwinn, L., Nguyen, A., Raab, R., Zanca, D., Eskofier, B.~M., Tenbrinck, D., and Burger, M.
\newblock Dynamically sampled nonlocal gradients for stronger adversarial attacks.
\newblock In \emph{2021 International Joint Conference on Neural Networks (IJCNN)}, pp.\  1--8. IEEE, 2021.

\bibitem[Schwinn et~al.(2023)Schwinn, Dobre, G{\"u}nnemann, and Gidel]{schwinn2023adversarial}
Schwinn, L., Dobre, D., G{\"u}nnemann, S., and Gidel, G.
\newblock Adversarial attacks and defenses in large language models: Old and new threats.
\newblock In \emph{NeurIPS 2023, ICBINB Workshop (Spotlight with Oral presentation)}, 2023.

\bibitem[Schwinn et~al.(2024)Schwinn, Dobre, Xhonneux, Gidel, and Gunnemann]{schwinn2024soft}
Schwinn, L., Dobre, D., Xhonneux, S., Gidel, G., and Gunnemann, S.
\newblock Soft prompt threats: Attacking safety alignment and unlearning in open-source llms through the embedding space.
\newblock \emph{arXiv preprint arXiv:2402.09063}, 2024.

\bibitem[Smith \& Topin(2018)Smith and Topin]{smith_super-convergence_2018}
Smith, L.~N. and Topin, N.
\newblock Super-convergence: {{Very}} fast training of neural networks using large learning rates.
\newblock \emph{ICLR 2018}, 2018.

\bibitem[Szegedy et~al.(2014)Szegedy, Zaremba, Sutskever, Bruna, Erhan, Goodfellow, and Fergus]{szegedy_intriguing_2014}
Szegedy, C., Zaremba, W., Sutskever, I., Bruna, J., Erhan, D., Goodfellow, I., and Fergus, R.
\newblock Intriguing properties of neural networks.
\newblock \emph{arXiv preprint arXiv:1312.6199}, 2014.

\bibitem[Toneva et~al.(2018)Toneva, Sordoni, Combes, Trischler, Bengio, and Gordon]{toneva2018empirical}
Toneva, M., Sordoni, A., Combes, R. T.~d., Trischler, A., Bengio, Y., and Gordon, G.~J.
\newblock An empirical study of example forgetting during deep neural network learning.
\newblock \emph{arXiv preprint arXiv:1812.05159}, 2018.

\bibitem[Wang et~al.(2020)Wang, Zou, Yi, Bailey, Ma, and Gu]{wang_improving_2020}
Wang, Y., Zou, D., Yi, J., Bailey, J., Ma, X., and Gu, Q.
\newblock Improving {{Adversarial Robustness Requires Revisiting Misclassified Examples}}.
\newblock \emph{ICLR 2020}, 2020.

\bibitem[Wang et~al.(2023)Wang, Pang, Du, Lin, Liu, and Yan]{wang_better_2023}
Wang, Z., Pang, T., Du, C., Lin, M., Liu, W., and Yan, S.
\newblock Better diffusion models further improve adversarial training.
\newblock \emph{PMLR 2023}, 2023.

\bibitem[Wong et~al.(2020)Wong, Rice, and Kolter]{wongFastBetterFree2020}
Wong, E., Rice, L., and Kolter, J.~Z.
\newblock Fast is better than free: {{Revisiting}} adversarial training.
\newblock \emph{ICLR 2020}, 2020.

\bibitem[Xhonneux et~al.(2024)Xhonneux, Sordoni, G{\"u}nnemann, Gidel, and Schwinn]{xhonneux2024efficient}
Xhonneux, S., Sordoni, A., G{\"u}nnemann, S., Gidel, G., and Schwinn, L.
\newblock Efficient adversarial training in llms with continuous attacks.
\newblock \emph{arXiv preprint arXiv:2405.15589}, 2024.

\bibitem[Xu et~al.(2023)Xu, Zhang, Liu, Sugiyama, and Kankanhalli]{xu_efficient_2023}
Xu, X., Zhang, J., Liu, F., Sugiyama, M., and Kankanhalli, M.
\newblock Efficient adversarial contrastive learning via robustness-aware coreset selection.
\newblock \emph{NeurIPS 2023}, 2023.

\bibitem[Yang et~al.(2023)Yang, Xie, Peng, Xu, Sun, and Li]{yangDatasetPruningReducing2023}
Yang, S., Xie, Z., Peng, H., Xu, M., Sun, M., and Li, P.
\newblock Dataset {{Pruning}}: {{Reducing Training Data}} by {{Examining Generalization Influence}}.
\newblock \emph{ICLR 2023}, 2023.

\bibitem[Zagoruyko \& Komodakis(2017)Zagoruyko and Komodakis]{zagoruyko_wide_2017}
Zagoruyko, S. and Komodakis, N.
\newblock Wide residual networks.
\newblock \emph{arXiv preprint arXiv:1605.07146}, 2017.

\bibitem[Zhang et~al.(2019)Zhang, Yu, Jiao, Xing, Ghaoui, and Jordan]{zhang_theoretically_2019}
Zhang, H., Yu, Y., Jiao, J., Xing, E.~P., Ghaoui, L.~E., and Jordan, M.~I.
\newblock Theoretically principled trade-off between robustness and accuracy.
\newblock \emph{PMLR 2019}, 2019.

\end{thebibliography}
    \bibliographystyle{icml2024}
    
    \newpage
    \appendix
    \onecolumn
    \section{Extrapolation Hyperparamters}
    The set of hyperparameter settings for the extrapolation of the data importance scores depends on the data importance metric and the attack type. The configuration used for our experiments is shown in \ref{tbl:hyperparameter}. For all experiments, selecting the source dataset as the combination of $S_{train}$ and the original CIFAR-10 dataset yielded the best results. In our tested settings, the DINOv2 model is never used as the encoding strategy. 
    \begin{table}[h]
    \caption{Hyperparameters selected for different configurations of the data importance extrapolation.}
    \label{tbl:hyperparameter}
    \vskip 0.15in
    \begin{center}
    \begin{small}
    \begin{sc}
        \begin{tabular}{llllll}
		\toprule[\heavyrulewidth]
		Attack                & Score & K  & Encoding & Distance Metric  & Training Data                             \\ \midrule[\lightrulewidth]
		\multirow{2}{*}{$L_\infty$} & DU    & 13 & SSCD     & Cosine Sim.    & \multirow{4}{*}{Normal+Synth.}  \\
							  & FP    & 12 & SSCD     & Euclidean Dist. & \\ \cmidrule(r){1-5} 
		\multirow{2}{*}{$\ell_2$}   & DU    & 35 & ResNet50 & Cosine Sim.    &  \\
							  & FP    & 24 & SSCD     & Euclidean Dist. &                                            \\ \bottomrule[\heavyrulewidth]   
        \end{tabular}
    \end{sc}
    \end{small}
    \end{center}
    \vskip -0.1in
    \end{table}

    


    \begin{acronym}
        \acro{APGD}{Auto-\ac{PGD}}
        \acro{CNN}{Convolutional Neural Network}
        \acro{DDPM}{Denoising Diffusion Probabilistic Model}
        \acro{DFT}{Discrete Fourier Transform}
        \acro{DL}{Deep Learning}
        \acro{DLR}{Difference of Logits Ratio}
        \acro{DNN}{Deep Neural Network}
        \acro{DU}{Dynamic Uncertainty}
        \acro{EDM}{Elucidating Diffusion Model}
        \acro{EDU}{Extrapolated \ac{DU}}
        \acro{EFP}{Extrapolated \ac{FP}}
        \acro{FAB}{Fast Adaptive Boundary}
        \acro{FFP}{Fundamental Frequency Pruning}
        \acro{FFT}{Fast Fourier Transform}
        \acro{FID}{Frechet Inception Distance}
        \acro{FP}{Frequency Pruning}
        \acro{GAN}{Generative Adversarial Network}
        \acro{GT}{Ground Truth}
        \acro{IDBH}{Improved Diversity and Balanced Hardness}
        \acro{KNN}{K-Nearest Neighbors}
        \acro{MAE}{Mean Absolute Error}
        \acro{MART}{Misclassification Aware adveRsarial Training}
        \acro{ML}{Machine Learning}
        \acro{MLP}{Multilayer Perceptron}
        \acro{NN}{Neural Network}
        \acro{PGD}{Projected Gradient Descent}
        \acro{ResNet}{Residual Neural Network}
        \acro{SGD}{Stochastic Gradient Descent}
        \acro{SOTA}{State Of The Art}
        \acro{SSCD}{Self Supervised Copy Detection}
        \acro{SVM}{Support Vector Machine}
        \acro{SWA}{Stochastic Weight Averaging}
        \acro{TRADES}{TRadeoff-inspired Adversarial DEfense via Surrogate-loss minimization}
        \acro{VAE}{Variational Autoencoder}
        \acro{ViT}{Vision Transformer}
        \acro{AA}{AutoAttack}
    \end{acronym}

    \end{document}